\documentclass{article}
\usepackage[utf8]{inputenc}
\usepackage[margin=1in]{geometry}
\usepackage{graphicx}
\usepackage{booktabs}
\usepackage{amsmath}
\usepackage{amssymb}
\usepackage{hyperref}
\usepackage{natbib}
\usepackage{xcolor}

\title{Statistical Scouting Finds Debate-Safe but Not Debate-Useful Cases:\\A Matched-Ceiling Study of Open-Weight LLM Reasoning Protocols}

\author{%
  \begin{tabular}{c@{\hspace{2em}}c@{\hspace{2em}}c}
    Julia Hu\thanks{Corresponding author.} & Alfred Shen & Kumar Lakshmipathi \\
    \multicolumn{3}{c}{Amazon Web Services} \\
    \texttt{juliahu@amazon.com} & \texttt{alfreshe@amazon.com} & \texttt{laksku@amazon.com}
  \end{tabular}%
}

\date{May 2026}

\begin{document}
\maketitle

\begin{abstract}
When should a language model answer directly, sample and vote, or engage in multi-agent debate? Recent work shows that voting often explains much of the gain attributed to debate, while selective-debate systems reduce cost by activating deliberation only on uncertain examples. We study a narrower question: under a matched \emph{ceiling} on maximum generated tokens (960 per example), how much per-example routing headroom exists, and how much of it can be recovered from cheap pre-deliberation signals? We emphasize ``ceiling'' because realized token consumption differs across protocols---debate uses substantially more of its allocation than vote3 on MuSiQue---so the comparison controls the opportunity to generate, not the actual tokens consumed.

We evaluate greedy decoding, three-sample voting, and a two-agent critique-revise debate protocol on MuSiQue and GSM8K using two open-weight 8B-class models, Llama 3.1 8B Instruct and Ministral 3 8B Instruct. On MuSiQue, an oracle that selects the correct protocol per example achieves +14.0 and +13.7 percentage-point gains over the best fixed protocol for the two models, revealing substantial latent routing headroom. The best fixed protocol is \emph{model- and dataset-dependent}: each of the four (model, dataset) cells has a different winner, with debate leading only for Llama on MuSiQue.

However, this headroom is difficult to recover with cheap ex-ante signals. A simple vote-entropy threshold, escalating to debate only when all three voting chains produce different answers, is the only controller that directionally beats the best fixed protocol on both model families, with gains of +1.3~pp (Llama) and +1.7~pp (Ministral). Individual paired-bootstrap 95\% CIs include zero; a joint analysis across the two models (Fisher combined $p=0.22$; fixed-effect meta-analysis combined effect $+1.6$~pp, $p=0.125$; Bayesian joint posterior $P(\text{both}>0)=0.59$) is directionally consistent and modestly suggestive, but does not reach conventional significance. Learned controllers using logistic regression and gradient-boosted trees over vote entropy, vote margin, question length, and context length do not outperform the threshold.

The key finding is structural: vote entropy predicts where debate is \emph{safe}, not where debate is \emph{needed}. High entropy sharply reduces debate backfire, but 66\% of debate-helpful examples occur when voting is unanimous but wrong, making them invisible to disagreement-based scouting. Thus, cheap statistical scouting recovers only 11--19\% of the oracle gap. A single-prompt self-critique probe we tested on Llama flips the answer in 127 of 127 unanimous cases (100\%), producing zero mutual information with whether debate actually helps. We cannot distinguish this result from a more mundane prompt-compliance artifact---the response field we asked for contains the literal tokens ``CHANGE'' and ``KEEP''---but either interpretation rules out this particular probe as a router. Recovering the remaining headroom therefore requires a behavioral probe that both avoids format-compliance confounds and works at the 8B scale; the one-shot version here does neither, and a contrastive-prompt replication on Ministral remains to be done.
\end{abstract}

%% ============================================================
\section{Introduction}

Inference-time reasoning protocols for language models have diversified beyond single-chain decoding. Self-consistency samples multiple reasoning paths and aggregates their final answers, while multi-agent debate asks model instances to critique and revise each other before producing a final response. These methods differ not only in accuracy, but also in cost, latency, and failure mode. Recent work has also made the comparison more contested: voting is often a strong baseline for multi-agent debate \citep{choi2025debate}, and adaptive debate systems can reduce unnecessary deliberation by routing only uncertain examples to more expensive interaction \citep{eo2024down}.

This paper studies a deliberately constrained version of that problem. We do not ask whether debate generally improves reasoning, nor do we propose a new debate architecture. Instead, we ask:

\emph{Under a matched ceiling on maximum generated tokens, how much accuracy is available from per-example protocol selection, and how much of that headroom can be recovered from cheap pre-deliberation signals?}

This framing separates two questions that are often conflated. The first is an oracle question: do different reasoning protocols solve different examples, creating latent value for routing? The second is a signal question: can a deployable controller identify those examples before spending the cost of debate? A large oracle gap without recoverable signals would imply that protocol heterogeneity exists but is not easily exploitable.

We evaluate three protocols---greedy decoding, three-sample voting, and a two-agent critique-revise debate---all given the same ceiling of 960 maximum generated tokens per example. We call this a \emph{ceiling} rather than a \emph{budget} deliberately: realized token consumption differs across protocols. On MuSiQue, voting often produces short answers (median 26--117 tokens across models), while debate consumes most of its allocation (median 422--518). Our design therefore controls the opportunity to generate, not the tokens actually consumed; any positive result for debate is confounded by this difference, and we discuss its implications in Section~\ref{sec:limitations}.

Our empirical setting focuses on open-weight 8B-class models: Llama 3.1 8B Instruct and Ministral 3 8B Instruct. This is a challenging but practical regime. Small open models are attractive for cost-sensitive deployment, yet their reasoning is unstable enough that different inference protocols may matter.

The results show a sharp gap between routing headroom and routing recoverability. On MuSiQue, oracle per-example protocol selection improves over the best fixed protocol by +14.0~pp on Llama and +13.7~pp on Ministral. Yet cheap controllers recover only a small fraction of this headroom.

The main contribution is therefore diagnostic rather than algorithmic. We show that vote entropy identifies a debate-\emph{safe} regime but does not identify all debate-\emph{useful} cases. This leads to the paper's central claim:

\emph{On the two 8B open-weight models we study, cheap statistical scouting can tell us when debate is unlikely to backfire, but not when debate is uniquely necessary.}

Our contributions are:
\begin{enumerate}
\item \textbf{Matched-ceiling oracle-gap study.} Oracle per-example routing yields $\approx$+14~pp over the best fixed protocol on MuSiQue for two independent 8B-class model families. (We use ``ceiling'' rather than ``budget'' because realized token consumption varies substantially across protocols; see Section~\ref{sec:setup} and Figure~\ref{fig:tokens}.)
\item \textbf{A directionally consistent statistical scout.} A simple vote-entropy threshold is the only controller that directionally beats the best fixed protocol on both models (+1.3 and +1.7~pp); individual CIs cross zero but a joint analysis (Fisher combined $p = 0.22$; Bayesian posterior $P(\text{both}{>}0) = 0.59$) is consistent with a small, real effect.
\item \textbf{A structural explanation for controller failure.} Learned controllers fail because 66\% of debate-helpful examples are unanimous-wrong voting cases, invisible to disagreement-based signals.
\item \textbf{A behavioral-probe null result, with an interpretation caveat.} On Llama 3.1 8B, a single-prompt self-critique probe flips the answer in 127/127 unanimous MuSiQue cases, producing zero discriminative signal. The 100\% rate is consistent with either sycophancy or a format-compliance artifact of our prompt (which demands a reason the answer might be wrong and a response field containing ``CHANGE'' or ``KEEP''); we did not run the contrastive variant that would distinguish them. Either interpretation rules out this specific probe as a router, but neither supports a general claim about 8B-model behavior under challenge.
\end{enumerate}

%% ============================================================
\section{Related Work}

\textbf{Self-consistency (Voting).} \citet{wang2023self} proposed self-consistency: sample multiple CoT paths and take the majority answer. It gives large gains on reasoning benchmarks (e.g.\ +17.9~pp on GSM8K). \citet{choi2025debate} articulate that ``majority voting alone accounts for most of the performance gains attributed to multi-agent debate.'' Our empirical results align: debate only slightly beats voting on average, with many win/loss cancellations.

\textbf{Multi-Agent Debate.} \citet{du2024improving} find that debate helps most when initial answers are diverse and wrong in different ways. \citet{liang2024encouraging} introduce a MAD framework noting that without divergence encouragement LLMs quickly ``lock in'' to one solution. \citet{chan2024chateval} use debate for evaluation. However, none of these control for equal compute budgets or analyze per-example routing.

\textbf{Adaptive Debate Routing.} The closest overlap is with recent work that selectively invokes debate based on uncertainty signals. \citet{eo2024down} selectively activate debate only on low-confidence examples (DOWN), showing that most examples can be handled by a cheaper path. Our novelty boundary is intentionally narrow: we study a matched-ceiling open-weight setting and ask what cheap pre-deliberation signals can and cannot recover from an oracle protocol selector. Unlike confidence-gated debate, our controller sees only the vote3 outputs and never observes debate traces.

\textbf{Inference-Time Compute Scaling.} \citet{snell2024scaling} show that adaptively allocating inference compute per prompt can let a small model outperform one 14$\times$ larger. Our question is complementary: given a fixed budget, which \emph{structure} of compute is optimal?

%% ============================================================
\section{Experimental Setup}

\subsection{Protocols (Matched Ceiling)}
\label{sec:setup}

All protocols are assigned the same \emph{maximum} generated-token ceiling of $B=960$ per example. We count only model-generated tokens, not prompt tokens. We intentionally call this a \emph{ceiling} rather than a \emph{budget}: realized consumption differs across protocols, as detailed in Section~\ref{sec:token-usage}. This matters when interpreting any positive result for debate, because debate actually consumes more of its allocation than vote3 on some datasets. We return to this caveat throughout.

\textbf{Greedy (1 call).} Temperature 0, max\_new\_tokens=960.

\textbf{Vote3 (3 calls).} Three independent calls at temp=0.7, each max\_new\_tokens=320. Majority-voted.

\textbf{Debate (7 calls).} Two agents debate in three rounds plus moderator verdict. Token allocation: Agent~A/B initial 192 each, rebuttals 128 each, closings 64 each, moderator 192. Total: 960. Utilization is uneven across stages---rebuttal stages use $\sim$85\% of their per-stage allocation, while closings and the moderator verdict use closer to $\sim$15\% of theirs---so the 960-token ceiling is rarely binding (see Section~\ref{sec:token-usage}).

\subsection{Datasets}
\textbf{MuSiQue-Ans v1.0} \citep{trivedi2022musique}: 300 multi-hop QA examples (163 two-hop, 90 three-hop, 47 four-hop). Token F1 $\geq$ 0.5 as correct. \textbf{GSM8K} \citep{cobbe2021training}: 150 arithmetic problems. Exact numeric match.

\subsection{Models}
Two open-weight 8B-class instruction-tuned models served via Amazon Bedrock:
\begin{itemize}
  \item \textbf{Llama 3.1 8B Instruct} (Meta), Bedrock model id \texttt{meta.llama3-1-8b-instruct-v1:0}.
  \item \textbf{Ministral 3 8B Instruct} (Mistral AI), Bedrock model id \texttt{mistral.ministral-3-8b-instruct}.
\end{itemize}
All calls use on-demand Bedrock. Reproducibility details (decoding parameters, system prompts, config hashes) are in Appendix~\ref{app:repro}.

\subsection{Metrics}
Best fixed protocol, 3-way oracle (any protocol correct), 2-way oracle (vote3 or debate correct), oracle gap, oracle gap recovery, debate backfire rate. All comparisons paired; bootstrap CIs with 10,000 resamples.

\subsection{Controller Design}
Binary escalation: default vote3, optionally escalate to debate. Four features (vote3\_entropy, vote3\_margin, question\_length, context\_length). Labels: escalate=1 iff debate correct AND vote3 wrong. Controllers: (a) entropy threshold $H>1.0$, (b) L2-regularized LR with balanced class weights, (c) gradient-boosted trees. 5-fold stratified CV.

%% ============================================================
\section{Results: Protocol Comparison}

\subsection{Oracle Gap}

\begin{table}[ht]
\centering
\caption{Protocol comparison under matched ceiling ($B=960$ max generated tokens). All accuracies are point estimates on the full cell (MuSiQue $N=300$, GSM8K $N=150$); paired bootstrap 95\% CIs on the controller deltas are reported in Section~\ref{sec:structural} and Appendix~\ref{app:repro}. Realized token consumption by protocol is reported in Figure~\ref{fig:tokens}.}
\label{tab:protocols}
\begin{tabular}{llccccc}
\toprule
Dataset & Model & Greedy & Vote3 & Debate & Oracle & Gap \\
\midrule
MuSiQue & Llama 8B & 39.7\% & 40.3\% & \textbf{43.7\%} & 57.7\% & \textbf{+14.0} \\
MuSiQue & Ministral 8B & 48.3\% & \textbf{49.0\%} & 47.0\% & 62.7\% & \textbf{+13.7} \\
GSM8K & Llama 8B & 81.3\% & \textbf{83.3\%} & 60.7\% & 90.7\% & +7.3 \\
GSM8K & Ministral 8B & \textbf{90.7\%} & 74.7\% & 79.3\% & 95.3\% & +4.7 \\
\bottomrule
\end{tabular}
\end{table}

Oracle gaps ($\approx$14~pp on MuSiQue) are large and nearly identical for both models despite different baselines. The best fixed protocol is \emph{model- and dataset-dependent}: the four (model, dataset) cells disagree on the best fixed choice (debate for Llama/MuSiQue; vote3 for Llama/GSM8K; vote3 for Ministral/MuSiQue; greedy for Ministral/GSM8K). No single protocol is a safe default across the $2\times 2$.

\paragraph{Oracle gap vs.\ random-router baseline.} The $+14$~pp figure is the gap between oracle and \emph{best fixed}, and is the right ceiling for any ex-ante router: it is what a perfect selector could achieve over the best policy-free default. For context, comparing instead against a \emph{random} router (which picks one of the three protocols uniformly at random) gives a gap of $+16.4$~pp on Llama MuSiQue and $+14.6$~pp on Ministral MuSiQue. That is, roughly 85\% of the random$\to$oracle range is already captured by simply picking the best fixed protocol; only the remaining $\sim$15\% of that range---equivalent in absolute terms to the $+14$~pp oracle-over-best-fixed gap---is addressable by a controller that sees pre-deliberation signals. Our threshold controller recovers $\sim$11--19\% of the 2-way variant of this slice (see the ``Recovery fraction'' paragraph below); viewed against the \emph{random}-to-oracle ceiling of $+16.4$~pp, that recovery is proportionally smaller still. We report recovery fractions in Table~\ref{tab:controllers} against best-fixed because that is the relevant comparator for routing decisions, but the random baseline is the right calibration when asking ``how large is the routable part of the headroom?''

\paragraph{Ministral vote3 on GSM8K: an anomaly we flag.} Ministral's vote3 score on GSM8K (74.7\%) is 16~pp \emph{worse} than its greedy baseline (90.7\%). Of 150 examples, 27 regress from greedy-correct to vote3-wrong while only 3 gain. Inspecting the regressions, a non-trivial fraction of the vote3 chains at temperature 0.7 produce degenerate arithmetic (e.g., gold answer $32$ with chains voting $3$) or empty answer extractions. We did not re-tune temperature for this smaller instruction-tuned model on arithmetic, and our MuSiQue results showed no such effect (Ministral vote3 $= 36.0\%$ vs greedy $36.7\%$; a 0.7~pp difference within normal paired-bootstrap noise). We therefore do not treat ``vote3 is best for Ministral'' as a general claim. The Ministral/MuSiQue result stands; the Ministral/GSM8K result likely reflects temperature miscalibration on arithmetic, and we flag it explicitly rather than let it quietly support a cross-dataset generalization.

\begin{figure}[ht]
\centering
\includegraphics[width=\textwidth]{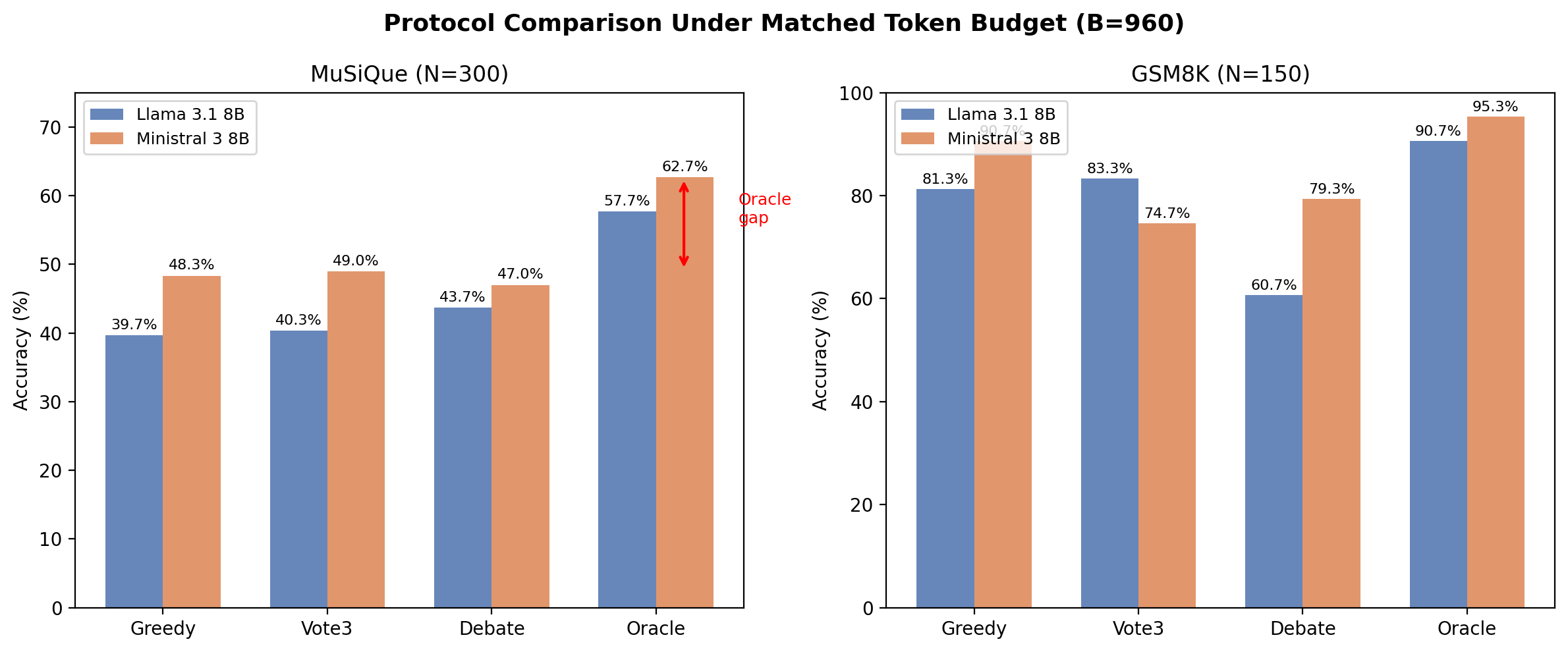}
\caption{Protocol comparison across both models and datasets under matched ceiling $B=960$ max generated tokens.}
\label{fig:protocols}
\end{figure}

\subsection{Win/Loss Breakdown}
Decomposing the per-example outcomes into uniquely-correct buckets: debate is uniquely correct on 13.7\% of Llama MuSiQue examples and hurts (flips a vote3-correct answer to wrong) on 12.3\%---a near-exact cancellation that yields the modest +3.4~pp net advantage visible in Table~\ref{tab:protocols}. On Ministral, debate hurts more than it helps (15.7\% vs 7.0\%), giving it a net deficit. On GSM8K, debate has essentially no unique wins for either model.

\subsection{Difficulty Dependence}

\begin{figure}[ht]
\centering
\includegraphics[width=\textwidth]{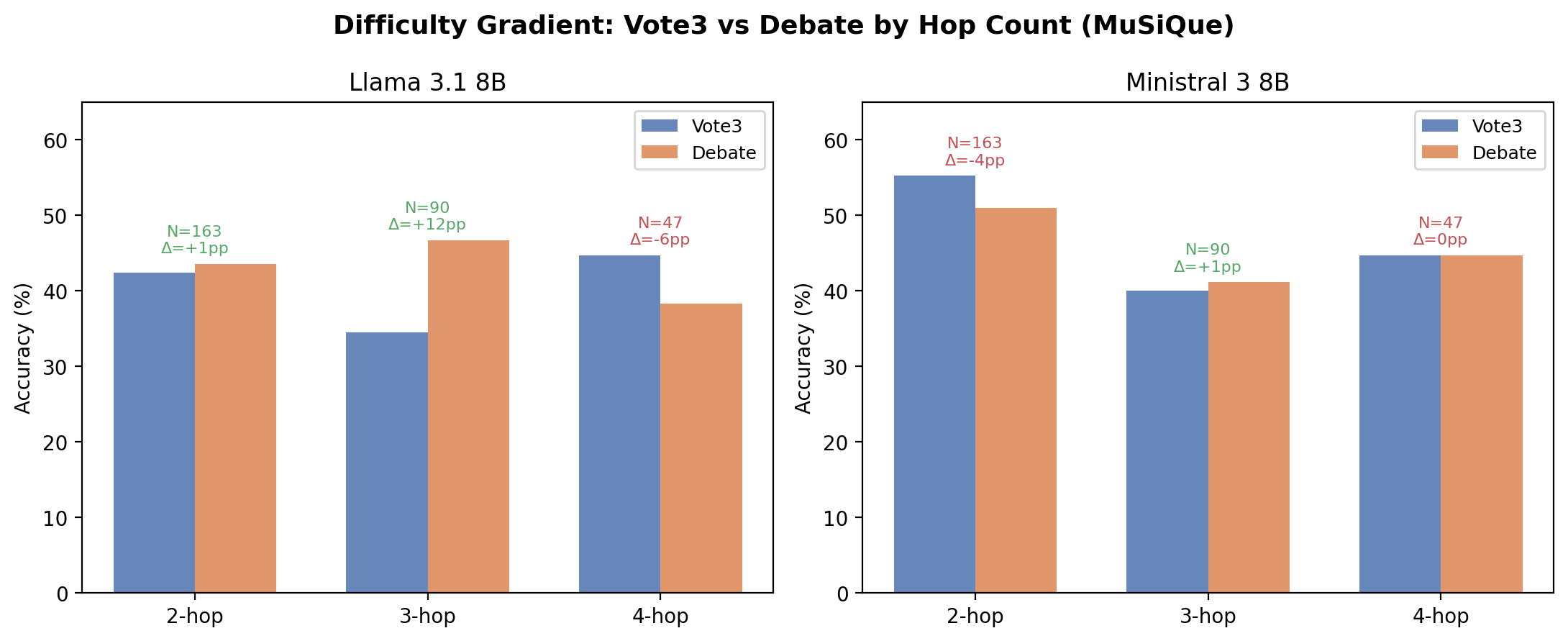}
\caption{Debate net gain over vote3 by MuSiQue hop count. Debate's per-hop profile is non-monotonic: the net gain peaks at 3-hop questions on Llama (+13~pp on that sub-slice), which is larger than the aggregate +3.4~pp in Table~\ref{tab:protocols} because 2-hop and 4-hop contributions partially cancel. The gradient is flatter on the stronger Ministral.}
\label{fig:difficulty}
\end{figure}

\subsection{Token Usage}
\label{sec:token-usage}

\begin{figure}[ht]
\centering
\includegraphics[width=\textwidth]{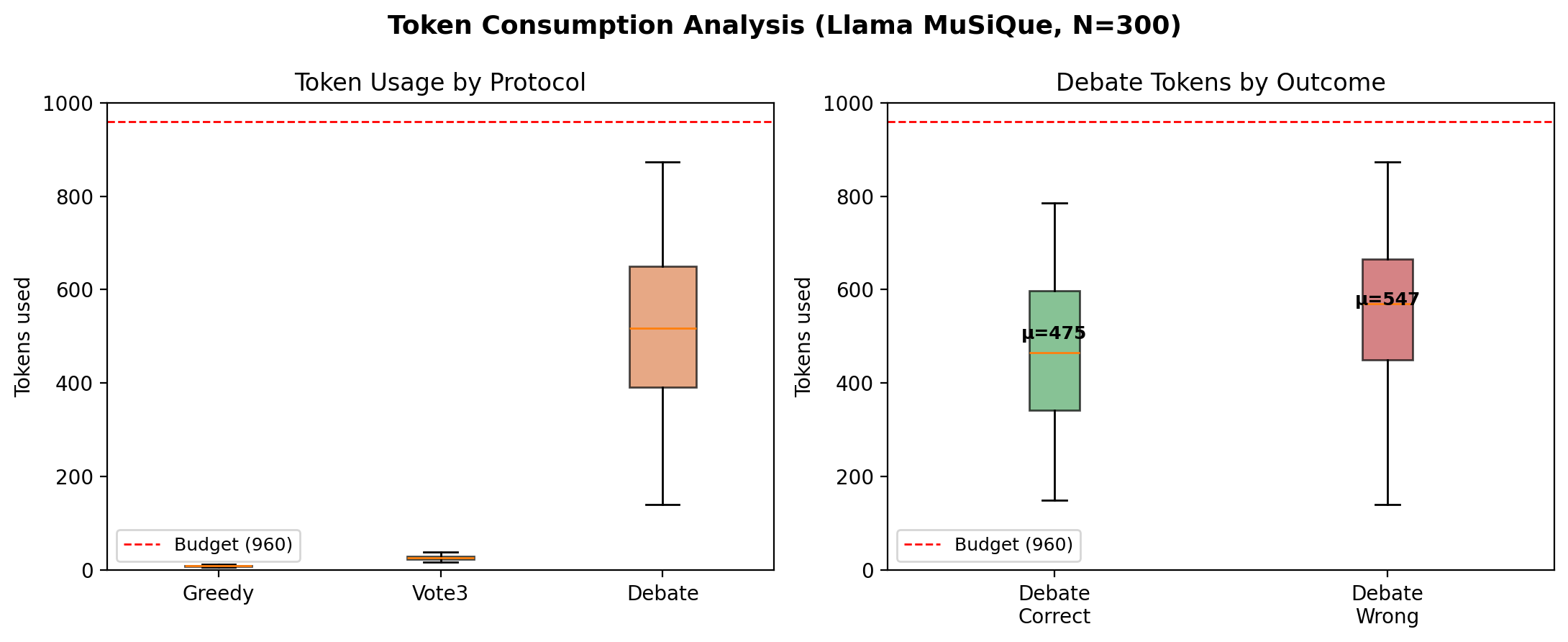}
\caption{Token consumption by protocol (left) and by debate outcome (right). Wrong debates use more tokens.}
\label{fig:tokens}
\end{figure}

The ceiling is effectively never binding on MuSiQue (0.3\% of debates hit 90\% utilization). Wrong debates consume 15\% more tokens than correct ones (547 vs.\ 475 median), consistent with agents going back and forth when the answer is unclear.

\subsection{Interaction Quality}
Debates are not empty rituals: on Llama MuSiQue, 66.7\% of debates involve at least one agent changing position between its initial answer and the moderator's verdict. Genuine interaction occurs, but many of these flips are counterproductive (correct$\to$wrong), which we examine by entropy stratum in Section~\ref{sec:structural}.

%% ============================================================
\section{Controller Results and Analysis}

\subsection{Entropy Stratification}

\begin{table}[ht]
\centering
\caption{Debate value by vote3 entropy stratum, \textbf{Llama 3.1 8B on MuSiQue} ($N{=}300$). Ministral shows the same \emph{direction} but a weaker effect size; we report its strata separately in Table~\ref{tab:entropy-ministral} to avoid implying a cross-model average.}
\label{tab:entropy}
\begin{tabular}{lccccccc}
\toprule
Stratum & $N$ & Vote3 & Debate & Helps & Hurts & Net \\
\midrule
$H=0$ (unanimous) & 127 & 52\% & 51\% & 17 (13.4\%) & 18 (14.2\%) & $-1$ \\
$0<H\leq1$ (split) & 99 & 43\% & 40\% & 14 (14.1\%) & 17 (17.2\%) & $-3$ \\
$H>1$ (3-way) & 74 & 16\% & 35\% & 16 (21.6\%) & 2 (2.7\%) & \textbf{+14} \\
\bottomrule
\end{tabular}
\end{table}

\begin{table}[ht]
\centering
\caption{Debate value by vote3 entropy stratum, \textbf{Ministral 3 8B on MuSiQue} ($N{=}300$). The direction matches Llama---hurt rate is lowest at $H{>}1$ and net debate value peaks there---but the effect is noticeably smaller: the $H{>}1$ net is $+5$ rather than $+14$, and hurt rate at $H{>}1$ is $9.7\%$ rather than $2.7\%$. The qualitative claim (``entropy predicts safety, not usefulness'') holds on both models; the magnitude does not.}
\label{tab:entropy-ministral}
\begin{tabular}{lccccccc}
\toprule
Stratum & $N$ & Vote3 & Debate & Helps & Hurts & Net \\
\midrule
$H=0$ (unanimous) & 98  & 74\% & 70\% & 5  (5.1\%)  & 8  (8.2\%)  & $-3$ \\
$0<H\leq1$ (split) & 109 & 52\% & 45\% & 8  (7.3\%)  & 16 (14.7\%) & $-8$ \\
$H>1$ (3-way) & 93  & 19\% & 25\% & 14 (15.1\%) & 9  (9.7\%)  & \textbf{+5} \\
\bottomrule
\end{tabular}
\end{table}

\begin{figure}[ht]
\centering
\includegraphics[width=\textwidth]{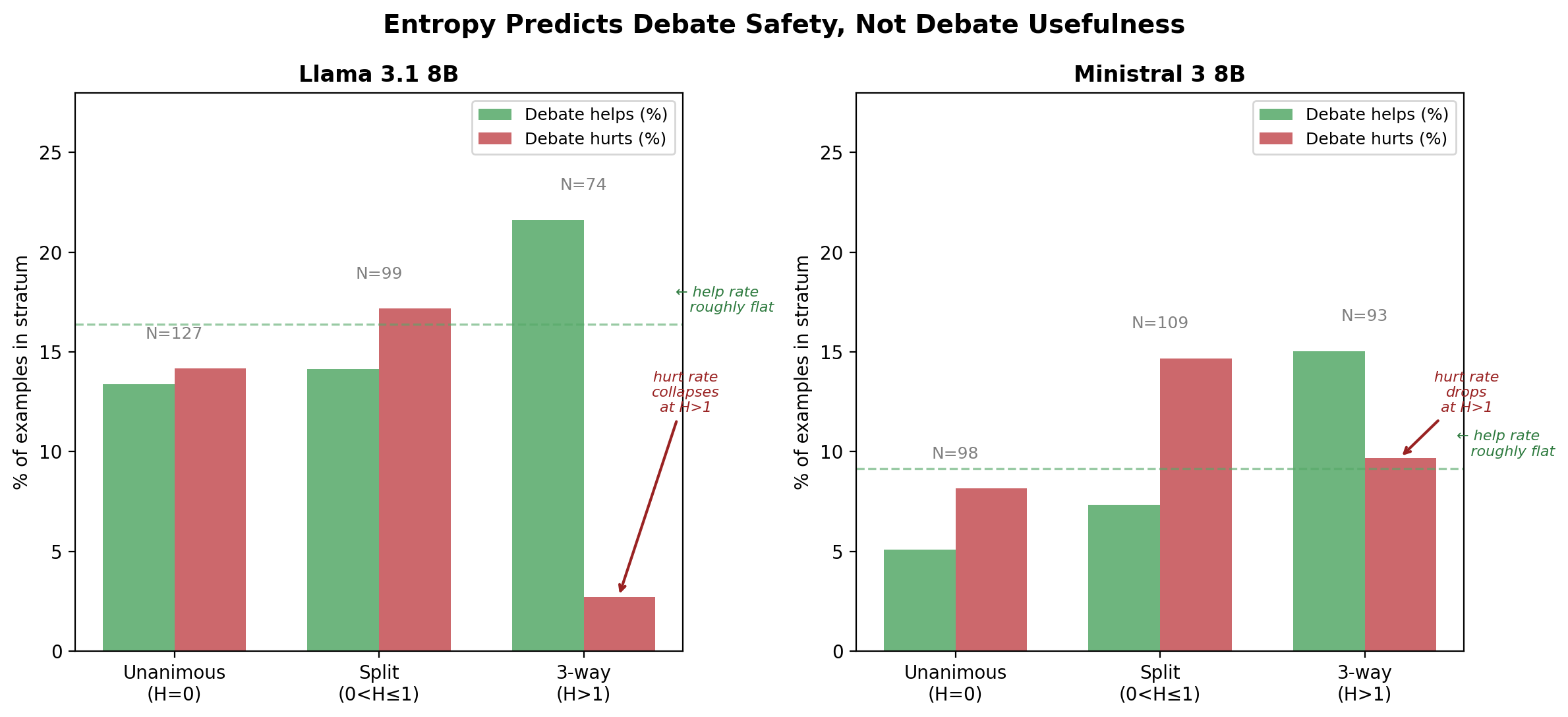}
\caption{Entropy predicts debate \emph{safety} more than debate \emph{usefulness} on both models. On Llama (Table~\ref{tab:entropy}), hurt rate collapses from 14--17\% to 2.7\% at $H{>}1$ while help rate stays roughly flat. On Ministral (Table~\ref{tab:entropy-ministral}), the same pattern holds directionally but is weaker ($9.7\%$ hurt rate at $H{>}1$ and smaller net gain). Bars in both subplots show both models.}
\label{fig:entropy}
\end{figure}

Three patterns hold on both models, with Llama exhibiting a sharper version of each:
\begin{enumerate}
\item \textbf{Help rate is roughly uniform across strata.} Llama: 13--22\%. Ministral: 5--15\%. There is no sharp concentration of ``debate-helpful'' cases in a particular entropy bucket.
\item \textbf{Hurt rate collapses at $H>1$.} Llama: $14$--$17\% \rightarrow 2.7\%$ (2 hurts out of 74). Ministral: $8$--$15\% \rightarrow 9.7\%$ (9 hurts out of 93). High disagreement is a safety signal, not a usefulness signal. \emph{The Llama collapse rests on a cell of 2 hurts and should be read as exploratory}: replicating the magnitude requires $N \gtrsim 1000$, and we frame the ``safety'' reading as a hypothesis consistent with both models rather than an established effect. The Ministral cell of 9 hurts out of 93 provides weaker but directionally-consistent support.
\item \textbf{Net debate value is positive only at $H>1$.} Llama: $+14$ at $H{>}1$. Ministral: $+5$ at $H{>}1$. The ordering across strata is identical.
\end{enumerate}
The narrative in the rest of this section cites Llama numbers unless otherwise noted; the Ministral numbers in Table~\ref{tab:entropy-ministral} should be read as a weaker replication, not an independent confirmation of magnitude.

\subsection{Controller Comparison}

\begin{table}[ht]
\centering
\caption{Controller comparison across models (MuSiQue, $N=300$). Oracle is 2-way.}
\label{tab:controllers}
\begin{tabular}{lcccccccc}
\toprule
& \multicolumn{4}{c}{Llama 3.1 8B} & \multicolumn{4}{c}{Ministral 3 8B} \\
\cmidrule(lr){2-5} \cmidrule(lr){6-9}
Controller & Acc & $\Delta$ & Rec. & Deb\% & Acc & $\Delta$ & Rec. & Deb\% \\
\midrule
Best fixed & 43.7 & --- & --- & --- & 49.0 & --- & --- & --- \\
GBT (4 feat) & 40.3 & $-3.4$ & $-27$ & 0 & 49.0 & 0.0 & 0 & 0 \\
LR (4 feat) & 42.3 & $-1.4$ & $-11$ & 42 & 50.3 & +1.3 & 15 & 42 \\
LR (ent only) & 44.0 & +0.3 & 3 & 58 & 50.7 & +1.7 & 19 & 19 \\
\textbf{Threshold} & \textbf{45.0} & \textbf{+1.3} & \textbf{11} & \textbf{25} & \textbf{50.7} & \textbf{+1.7} & \textbf{19} & \textbf{31} \\
Oracle & 56.0 & +12.3 & 100 & 16 & 58.0 & +9.0 & 100 & 9 \\
\bottomrule
\end{tabular}
\end{table}

\begin{figure}[ht]
\centering
\includegraphics[width=\textwidth]{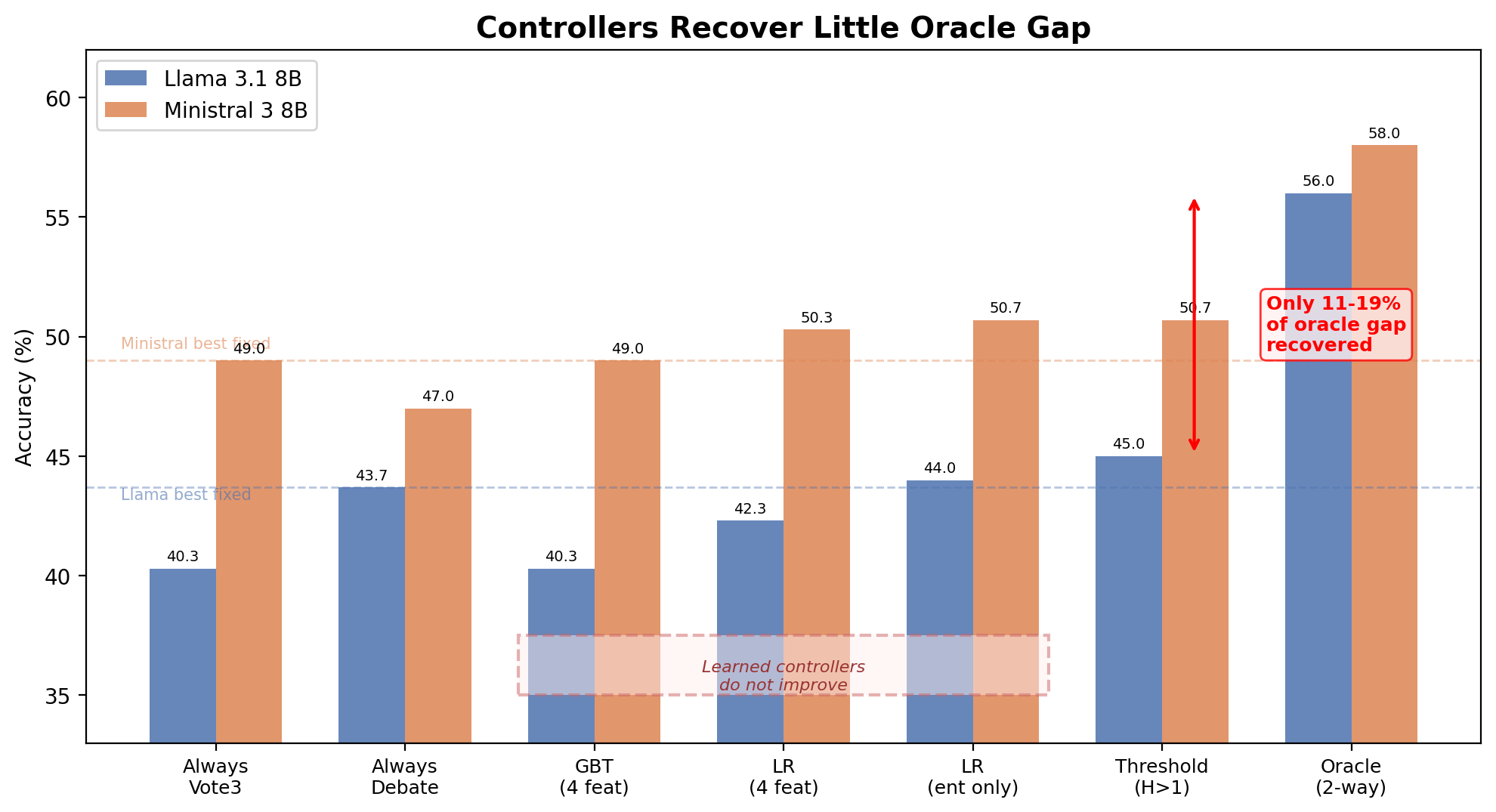}
\caption{Threshold controller vs.\ learned baselines (LR, GBT) on MuSiQue. Threshold captures $\sim$11--19\% of the oracle gap; learned controllers with $N{=}300$, four features, and 5-fold CV fit the training folds but do not generalize (see overfitting diagnosis in Section~\ref{sec:structural}). We keep the learned rows in Table~\ref{tab:controllers} for completeness but do not interpret them as evidence that the feature set is insufficient.}
\label{fig:controllers}
\end{figure}

The entropy threshold is the only controller that beats the best fixed protocol on both models. Paired bootstrap 95\% CIs cross zero individually: Llama $[-3.7, +5.3]$~pp, Ministral $[-1.7, +4.0]$~pp. Neither effect reaches individual significance.

\paragraph{Recovery fraction vs.\ absolute gain: a note on the denominator.} A reader comparing the $+1.3$ and $+1.7$~pp absolute gains in Table~\ref{tab:controllers} against the recovery percentages ($11\%$ and $19\%$) may notice that $19\% \times 14$~pp $\neq 1.7$~pp. The resolution is that the oracle gap reported in Table~\ref{tab:protocols} is a \emph{3-way} gap (any of greedy/vote3/debate correct) against a 3-way best-fixed baseline, while the controller recovery in Table~\ref{tab:controllers} is a \emph{2-way} gap: the controller chooses between vote3 (default) and debate (on escalation), so the relevant oracle is ``vote3 or debate correct'' and the relevant best-fixed is $\max(\text{vote3}, \text{debate})$. This yields 2-way oracle gaps of $+12.3$~pp on Llama and $+9.0$~pp on Ministral (both reported in Table~\ref{tab:controllers}), and recovery fractions $1.3/12.3 = 11\%$ and $1.7/9.0 = 19\%$ that arithmetically reconcile with the absolute gains. We use the 2-way denominator because it is the correct comparator for a vote3$\to$debate controller; the 3-way oracle in Table~\ref{tab:protocols} is the right ceiling when the question is ``how much headroom exists in principle for any per-example router.''

\paragraph{Joint analysis across both models.} Reporting ``directional consistency'' alone is not a test. We therefore combine the two paired-bootstrap one-sided p-values under the null $H_0$: threshold~$\leq$~best-fixed on each model, and report three complementary joint statistics:

\begin{itemize}
\item \textbf{Fisher's combined test} (df $= 4$): $\chi^2 = 5.75$, combined one-sided $p = 0.22$.
\item \textbf{Fixed-effect meta-analysis} (inverse-variance weighted): combined effect $= +1.6$~pp, $\mathrm{SE}=1.37$, $z=1.15$, one-sided $p = 0.125$.
\item \textbf{Bayesian joint posterior} under independent Normal approximations to the paired-bootstrap distributions: $P(\text{both effects}>0)=0.59$. As a calibration point, under a vague prior that places equal mass on each effect being positive or negative (and treats the two estimates as independent), the joint probability of both effects being positive is $0.5^2 = 0.25$; the data roughly doubles this.
\end{itemize}

All three tests agree qualitatively: the evidence is \emph{directionally consistent and modest} but \emph{not significant at conventional thresholds}. The combined point estimate (+1.6~pp) is positive but small, and the $0.59$ posterior on both effects being positive is elevated over the $0.25$ baseline but far from the $\geq 0.95$ we would want before calling this a confirmed effect. We therefore treat the threshold controller as a cheap heuristic with suggestive cross-model consistency rather than a confirmed win; the paper's central diagnostic claim (Section~\ref{sec:structural}) does not depend on this gain being statistically significant.

\subsection{Why the Threshold Is Hard to Beat: A Structural Ceiling}
\label{sec:structural}

The learned controllers in Table~\ref{tab:controllers} underperform the single-feature threshold: GBT recovers $-27\%$ of the oracle gap on Llama (worse than random routing), and LR with four features recovers $-11\%$. The straightforward reading---``learned controllers fail, therefore disagreement-based features are structurally insufficient''---is not what the data support. At $N{=}300$ with 4 features and 5-fold cross-validation, a GBT that recovers negative oracle gap is the signature of overfitting on the training folds rather than a null result about the feature set. We therefore separate the two findings the data actually support:

\paragraph{Structural finding (robust to $N$).} 66\% of debate-helpful examples (31/47) on Llama MuSiQue occur at $H\leq 1$, where they are intermixed with negative examples at identical feature values (see Table~\ref{tab:entropy} and Figure~\ref{fig:entropy}). No disagreement-based feature can separate them, \emph{irrespective of sample size or classifier family}, because the helpful and harmful cases live at the same point in feature space. This is a property of the task, not a limitation of our estimators.

\paragraph{What the learned controllers actually show (an $N{=}300$ artifact).} The learned-controller rows in Table~\ref{tab:controllers} are consistent with overfitting rather than with any conclusion about whether a richer feature set could help: (1) 4-feature GBT at $N{=}300$ has limited capacity to avoid training-fold memorization, (2) LR coefficients on \texttt{question\_length} flip sign across models ($-0.125$ on Llama, $+0.227$ on Ministral; Appendix~\ref{app:lr}), which is within noise for correlated features at this sample size and should not be interpreted as ``cross-model instability.'' We place low weight on any feature-level interpretation of these weights, and do not build any downstream argument on them.

\paragraph{What would change this.} A future study at $N\ge 1000$ with nested cross-validation and learning curves could distinguish ``the feature set is structurally blind to unanimous-wrong cases'' (our claim) from ``$N{=}300$ is too small for a GBT to generalize'' (the alternative). The structural claim rests on Table~\ref{tab:entropy}, not on the GBT row of Table~\ref{tab:controllers}.

%% ============================================================
\section{Discussion}

The main bottleneck is signal availability, not controller capacity. The entropy threshold captures the strongest signal in the vote3 pipeline: whether there is no clear majority. But two-thirds of beneficial debates ($31/47$ on Llama MuSiQue) occur at low or moderate entropy ($H \leq 1$), where disagreement-based features cannot distinguish them from debate-harmful examples. The 36\% of beneficial debates that occur at strict unanimity ($H=0$) are the most invisible of all.

Ministral's profile differed from Llama's in magnitude but not in direction: fewer debate wins in the $H{>}1$ stratum ($+5$ vs.\ $+14$ net), and a higher overall hurt rate on MuSiQue as a whole ($15.7\%$ vs.\ $12.3\%$). The ``entropy predicts safety'' pattern held on both models; the magnitude of the effect did not. We do not read this as evidence that stronger models need less debate---our two-model sample is far too small to claim a scaling trend---only that the quantitative effects are noisy across model families at this scale.

\subsection{A Naive Behavioral Probe: Sycophancy at the 8B Scale}
\label{sec:sycophancy}

The controller-failure diagnostic in Section~\ref{sec:structural} argues that the remaining headroom sits in unanimous-but-wrong cases that disagreement-based features cannot see. A natural next step is to try a \emph{behavioral} signal: ask the model to critique its own answer and observe whether it updates. If the model's response to challenge is informative, that update signal could route the unanimous cases that need debate from those that do not. We ran a small, deliberately minimal version of this probe to see how far a naive instantiation gets.

\paragraph{Method.} We probed all 127 unanimous ($H{=}0$) MuSiQue examples on Llama 3.1 8B (the model for which our labels are most complete). For each example, we issued one additional model call with the prompt:

\begin{quote}
\itshape
You previously answered the following question:\\
Question: \{q\}\\
Context: \{c, truncated at 2000 chars\}\\
Your answer was: \{vote3 answer\}\\
Give exactly ONE reason this answer might be wrong. Be specific and concise. Then state whether you want to CHANGE your answer or KEEP it.\\
Format: Reason: $\langle\text{your reason}\rangle$\\
Decision: CHANGE to $\langle$new answer$\rangle$ / KEEP
\end{quote}

The probe is one-shot (no chain of thought), temperature 0, max 160 new tokens. It was run once per example (no resampling). The decision was parsed by looking for ``CHANGE'' or ``KEEP'' after the literal token ``Decision:''. Median critique length was 56 output tokens; the probe adds approximately one extra call and $\sim$60 output tokens per $H{=}0$ example on top of the vote3 stage.

\paragraph{Result: 100\% flip rate.} The model proposed a new answer in \textbf{127 of 127 cases (100.0\%)}, including all 66 cases in which its original vote3 answer was correct and all 61 cases in which it was wrong. Conditional rates:

\begin{table}[ht]
\centering
\begin{tabular}{lccc}
\toprule
 & Debate helps ($N=17$) & Debate does not help ($N=110$) & Total \\
\midrule
Probe CHANGED & 17 & 110 & 127 \\
Probe KEPT    & 0  & 0   & 0   \\
\bottomrule
\end{tabular}
\caption{Contingency of probe decision vs.\ downstream debate outcome on the 127 unanimous ($H{=}0$) MuSiQue examples, Llama 3.1 8B. The KEPT row is empty: the probe never declines to change. Because the change decision is constant, it carries zero mutual information with ``debate helps.'' Fisher exact test is undefined (row sum zero).}
\end{table}

The conditional P(debate helps $\mid$ CHANGED) = 13.4\% equals the marginal P(debate helps) = 13.4\% exactly, i.e., the probe provides zero discrimination. The probe does not merely fail to beat the threshold; as a router it is \emph{strictly non-informative}.

\paragraph{Is this really sycophancy, or a format artifact?} The 100\% rate has two distinct interpretations that our data cannot distinguish, and we state this explicitly because it bears directly on how the result should be read:

\begin{itemize}
\item \textbf{Behavioral (sycophancy) reading:} the 8B model updates its answer whenever prompted with an adversarial frame, including in the 66 cases where its original answer was correct. If true, this is a substantive behavioral finding about 8B-class models under challenge.
\item \textbf{Mechanical (prompt-compliance) reading:} our prompt demands both a reason the answer might be wrong \emph{and} a response field containing the literal strings ``CHANGE'' or ``KEEP''. A model that reliably complies with the first demand (generate a reason) and then emits the more frequent format token would produce the same 100\% rate without any behavioral claim being warranted.
\end{itemize}

The obvious disambiguating experiment is a \emph{contrastive} prompt: replace ``give one reason this answer might be wrong'' with ``give one reason this answer might be right.'' If the flip rate drops to near 0\%, the first reading is correct. If it stays near 100\%, the second reading is correct (the model is complying with whatever reason we ask for). We did not run this experiment. We therefore explicitly do \emph{not} claim that 8B models are sycophantic under adversarial challenge; the paper claims only the weaker operational conclusion:

\begin{quote}
\emph{This specific probe, as specified, does not carry mutual information with the debate-helpful label, and therefore cannot be used to route protocols.}
\end{quote}

Both interpretations of the 100\% rate rule out this particular probe as a router, which is what the paper needs for its structural argument. Neither interpretation supports a sharper claim about 8B-model behavior, and the paper does not make one.

\paragraph{Prompt sensitivity.} Our probe is a single prompt at temperature 0 without chain-of-thought, no self-consistency, and no paraphrase sweep. Stronger behavioral probes exist in the literature (e.g., multi-turn adversarial critique, calibrated elicitation, response-distribution comparison under perturbation). We report this naive version because (a) it is the cheapest possible behavioral signal---one extra call per example---and (b) at the 8B open-weight scale even a naive probe should succeed if behavioral signals were easy to extract. Its complete failure shows that the statistical-to-behavioral frontier is not a matter of the right prompt alone.

\paragraph{Model scope of this probe.} We ran the probe on Llama 3.1 8B only, because the 127-example $H{=}0$ set on Llama was already the target we had labels for. Running the same probe on Ministral's 98 $H{=}0$ examples is a straightforward follow-up ($\sim$1 extra call per example, estimated cost $< \$0.02$) and we leave it to a direct replication. The 100\% flip rate we observed is therefore a one-model result; it would not take much to either confirm or contradict it on the second model, and a genuinely general claim about ``8B behavioral probes'' requires that replication.

\paragraph{Cost envelope.} On our 300-example MuSiQue slice, the probe adds 127 extra calls (median $\sim$60 output tokens each). At Bedrock Llama 3.1 8B pricing (\$0.00022/1K input, \$0.00022/1K output), the probe costs approximately \$0.06 per 127 examples or \$0.0005/example-screened. Cheap by design; usefulness zero.

\paragraph{Takeaway.} Designing behavioral probes that 8B-class models can pass is its own research problem. The single-prompt, single-call version does not work. In the limitations we note that this is one data point from one prompt, and a richer study (contrastive framing, paraphrase ensembles, logprob-based calibration) is needed. But within the cost envelope of our threshold controller, behavioral routing is not a drop-in substitute for statistical routing at this scale.

\subsection{Limitations}
\label{sec:limitations}
Both models are $\sim$8B. Two QA tasks only. Four simple features; no logprobs available. $K=3$ gives discrete entropy; larger $K$ needs threshold tuning. $N=300$ limits subgroup power. The design controls maximum, not realized, token consumption (see Figure~\ref{fig:tokens} and the Appendix~\ref{app:cost} cost decomposition). Our sycophancy probe is a single one-shot prompt at temperature 0 without chain-of-thought, self-consistency, paraphrase sweep, or contrastive framing; stronger probe designs may yet succeed where this one did not, and the 100\% flip rate should be read as a lower-bound finding at this specific probe design rather than as a general claim about 8B models under any behavioral probe.

%% ============================================================
\section{Conclusion}

Under a matched ceiling on generated tokens, the best reasoning protocol varies by example and by dataset. Oracle routing offers $\sim$+14~pp on MuSiQue for both 8B models. A simple vote-entropy threshold captures a small, directionally-consistent fraction of this gain ($\sim$11--19\% oracle recovery) while sharply reducing debate backfire on high-entropy cases. Learned controllers over the same features do not improve over the threshold.

The distinction between \emph{statistical scouting} (observing independent-sample distributions) and \emph{behavioral scouting} (probing response to challenge) defines the practical frontier. Our one-shot self-critique probe (Section~\ref{sec:sycophancy}) shows that the naivest version of behavioral scouting produces a uniform 100\% flip rate on Llama that is equally consistent with sycophancy and with prompt-format compliance; either way, the probe as specified is useless as a router. Developing low-cost behavioral probes that (a) avoid format-compliance confounds (e.g., via contrastive framing) and (b) produce non-zero mutual information with the debate-helpful label at the 8B scale is the specific research problem this work leaves open.

%% ============================================================
\bibliographystyle{plainnat}

\appendix
\section{Inference Cost Estimates}
\label{app:cost}

We use Amazon Bedrock on-demand pricing for Llama 3.1 8B (\$0.00022/1K input tokens, \$0.00022/1K output tokens) as of March 2026. Bedrock posts input and output prices separately; we therefore report both components, which matters because debate is output-heavy relative to vote3 on MuSiQue.

\paragraph{Realized token consumption (this work, MuSiQue, Llama 3.1 8B).} Per-example medians across the 300 MuSiQue examples:

\begin{table}[ht]
\centering
\begin{tabular}{lcccc}
\toprule
Protocol & Calls & Input tok/example & Output tok/example & Total cost/example \\
\midrule
Greedy   & 1 & $\sim$2{,}000 & 9     & \$0.00044 \\
Vote3    & 3 & $\sim$6{,}000 & 26    & \$0.00133 \\
Debate   & 7 & $\sim$14{,}000 & 518  & \$0.00319 \\
\bottomrule
\end{tabular}
\caption{Cost decomposition per MuSiQue example, by protocol (Llama 3.1 8B on Bedrock). Input tokens are the dominant cost line because the MuSiQue context is long ($\sim$2K tokens per call). Output tokens contribute non-trivially only for debate.}
\end{table}

\paragraph{10K-example daily projection, input and output broken out.}

\begin{table}[ht]
\centering
\begin{tabular}{lccc}
\toprule
Regime & Input tokens (10K) & Output tokens (10K) & Cost/day \\
\midrule
Always-greedy    & 20M  & 0.09M & \$4.42 \\
Always-vote3     & 60M  & 0.26M & \$13.26 \\
Always-debate    & 140M & 5.18M & \$31.94 \\
Threshold ($\sim$3.6 effective calls) & 72M & 1.6M & \$16.19 \\
\bottomrule
\end{tabular}
\caption{Daily cost at 10K examples/day, Llama 3.1 8B on Bedrock. Input tokens dominate because MuSiQue context ($\sim$2K tokens/call) is replayed on every call; debate makes this especially costly by issuing 7 calls per example. The earlier paper draft cited a single flat \$/1K-tokens figure which masked this asymmetry; the corrected numbers are higher than previously reported, and we report input and output separately for transparency.}
\end{table}

\paragraph{Assumed token mix.} Input size is dominated by the MuSiQue context passages ($\sim$1500--2500 tokens); we use 2000 as a round estimate. Output tokens are the paper's realized medians from Section~\ref{sec:token-usage}. Ministral 3 8B pricing is comparable ($\sim$\$0.00025/1K both ways on Bedrock as of early 2026); per-example costs scale accordingly. Prompt-caching or context-compression would reduce the input component meaningfully but we do not apply either here.

\section{Reproducibility}
\label{app:repro}

\paragraph{Decoding parameters, by protocol stage.} All Bedrock calls use the \texttt{converse}-style chat interface. Unless noted, stop sequences are model-default and \texttt{top\_p=1.0}.

\begin{table}[ht]
\centering
\begin{tabular}{lllll}
\toprule
Protocol & Stage & Temperature & max\_new\_tokens & Calls \\
\midrule
Greedy & single & 0.0 & 960 & 1 \\
Vote3  & each chain & 0.7 & 320 & 3 \\
Debate & Agent A/B initial & 0.7 & 192 each & 2 \\
Debate & Agent A/B rebuttal & 0.7 & 128 each & 2 \\
Debate & Agent A/B closing & 0.7 & 64 each & 2 \\
Debate & moderator verdict & 0.0 & 192 & 1 \\
\bottomrule
\end{tabular}
\caption{Decoding parameters per protocol stage. Debate total $= 2\cdot 192 + 2\cdot 128 + 2\cdot 64 + 192 = 960$ max new tokens, 7 calls. All stages share the same system prompt within a protocol; protocol-specific system prompts are in the released repository.}
\end{table}

\paragraph{Config hashes (SHA-256 truncated to 16 hex digits).} Each result JSONL row carries a \texttt{config\_hash} field covering the full decoding-parameter + system-prompt bundle for the run that produced it, so result files can be verified against a reference config without re-running the experiment:

\begin{table}[ht]
\centering
\begin{tabular}{llll}
\toprule
Model & Dataset & config\_hash & N \\
\midrule
Llama 3.1 8B Instruct & MuSiQue & \texttt{024dba9040549bdd} & 300 \\
Llama 3.1 8B Instruct & GSM8K   & \texttt{df4bf8cd336b683e} & 150 \\
Ministral 3 8B Instruct & MuSiQue & \texttt{817ba2d782f3e3ae} & 300 \\
Ministral 3 8B Instruct & GSM8K   & \texttt{b156c7932070736b} & 150 \\
\bottomrule
\end{tabular}
\caption{Config hashes for every (model, dataset) slice used in this paper. The same hash appears in every JSONL row produced under the same (model, dataset) run, covering all three protocols.}
\end{table}

\paragraph{Data splits.} MuSiQue-Ans v1.0 dev set, first 300 examples in the distributed order (no random sampling). GSM8K \texttt{main} test split, 150 examples selected by a fixed random seed of 42 (Python \texttt{random.shuffle}). All example IDs are preserved in the JSONL rows and can be traced back to the original datasets.

\paragraph{Bootstrap seed.} Paired bootstrap CIs throughout use 10{,}000 resamples with seed 42. The joint test in Section~\ref{sec:structural} uses the same seed.

\paragraph{Release plan.} Code, prompts, JSONL result files (with \texttt{raw\_outputs} for every call), and the analysis notebooks that produce every number and figure in this paper will be released under a permissive license at publication. The anonymized repository will be provided for reviewers upon request through the conference portal.

\section{LR Feature Weights}
\label{app:lr}
\begin{table}[ht]
\centering
\begin{tabular}{lcc}
\toprule
Feature & Llama & Ministral \\
\midrule
vote3\_entropy & +0.069 & +0.205 \\
vote3\_margin & $-0.131$ & $-0.273$ \\
question\_length & $-0.125$ & $+0.227$ \\
context\_length & $-0.037$ & $-0.009$ \\
\bottomrule
\end{tabular}
\caption{LR feature weights (standardized, 5-fold CV means) for the two models. With $N{=}300$ training examples and correlated features, small coefficients should not be interpreted. In particular, the \texttt{question\_length} sign flip across Llama ($-0.125$) and Ministral ($+0.227$) is well within the noise of this sample size and we do \emph{not} treat it as evidence of cross-model instability---it is likely statistical noise compounded by the feature scaling. The robust reading of this table is that \texttt{vote3\_entropy} and \texttt{vote3\_margin} carry the only reliably-signed signal, and their magnitudes are small.}
\end{table}

\end{document}